\documentclass[11pt,a4paper]{article}
\usepackage{times,latexsym}
\usepackage{url}
\usepackage[T1]{fontenc}
\usepackage{multirow}
\usepackage{amssymb}
\usepackage{subfigure}
\usepackage{amsmath}
\usepackage{float}
\usepackage{graphicx}
\usepackage{makecell}
\usepackage[normalem]{ulem}
\useunder{\uline}{\ul}{}

\makeatletter
\renewcommand{\maketag@@@}[1]{\hbox{\m@th\normalsize\normalfont#1}}%
\makeatother

\usepackage[acceptedWithA]{tacl2021v1}

\usepackage{xspace,mfirstuc,tabulary}

\newif\iftaclinstructions
\taclinstructionsfalse 
\iftaclinstructions
\renewcommand{\confidential}{}
\renewcommand{\anonsubtext}{(No author info supplied here, for consistency with
TACL-submission anonymization requirements)}
\newcommand{\instr}
\fi

\iftaclpubformat 

\else

\fi

\title{A Perplexity and Menger Curvature-Based Approach for Similarity Evaluation of Large Language Models}

\author{
Yuantao Zhang\textsuperscript{1,2},
~Zhankui Yang\textsuperscript{1}$\Thanks{Corresponding Author}$\\
\textsuperscript{1}National Supercomputing Center in Shenzhen
~\textsuperscript{2}National University of Singapore
}

\date{}

\begin{document}
\maketitle
\begin{abstract}
The rise of Large Language Models (LLMs) has brought about concerns regarding copyright infringement and unethical practices in data and model usage. For instance, slight modifications to existing LLMs may be used to falsely claim the development of new models, leading to issues of model copying and violations of ownership rights. This paper addresses these challenges by introducing a novel metric for quantifying LLM similarity, which leverages perplexity curves and differences in Menger curvature. Comprehensive experiments validate the performance of our methodology, demonstrating its superiority over baseline methods and its ability to generalize across diverse models and domains. Furthermore, we highlight the capability of our approach in detecting model replication through simulations, emphasizing its potential to preserve the originality and integrity of LLMs. Code is available at \url{https://github.com/zyttt-coder/LLM_similarity}.
\end{abstract}

\section{Introduction}
\label{sec1}
In the past year, the rapid development of Large Language Models (LLMs) and their wide application have become a hot spot in different domains. Although LLMs provide a more convenient way to acquire knowledge and solve problems, they also bring about some issues. Companies and organizations have begun to exploit LLMs for profit by engaging in unethical practices such as directly copying model structures and codes, and violating open-source licenses. For instance, the Yi-34B model developed by Chinese 01-ai company uses exactly Llama's architecture except for two tensors renamed\footnote{\href{https://huggingface.co/01-ai/Yi-34B/discussions/11}{hf.co/01-ai/Yi-34B/discussions/11}}. Additionally, there are cases where proprietary LLMs are rebranded with slight modifications, such as adding noise, and falsely represented as original creations. Recent studies have revealed that the Llama3-V project code from Stanford team is a reformulation of the MiniCPM-Llama3-V2.5 \cite{xu2024llava-uhd,yu2024rlaifv}, and the behavior of the Llama3-V model is very similar to a noised version of the MiniCPM-Llama3-V2.5 checkpoint. Besides, methods can be used to distill the knowledge of a LLM in specific areas into other LLMs \cite{xu2024survey}. Without explicit clarifications, such actions may potentially breach the model's user policies. While distilled models and the original LLM have similarity in the distilled areas, their performance could diverge in other domains, making detection more difficult. The practices mentioned above significantly undermine companies' exclusive rights to their own products, highlighting the need for effective approaches to measure the similarity between LLMs and uncover these activities.

\begin{figure}[t]
\centering
\includegraphics[width=1.0\columnwidth]{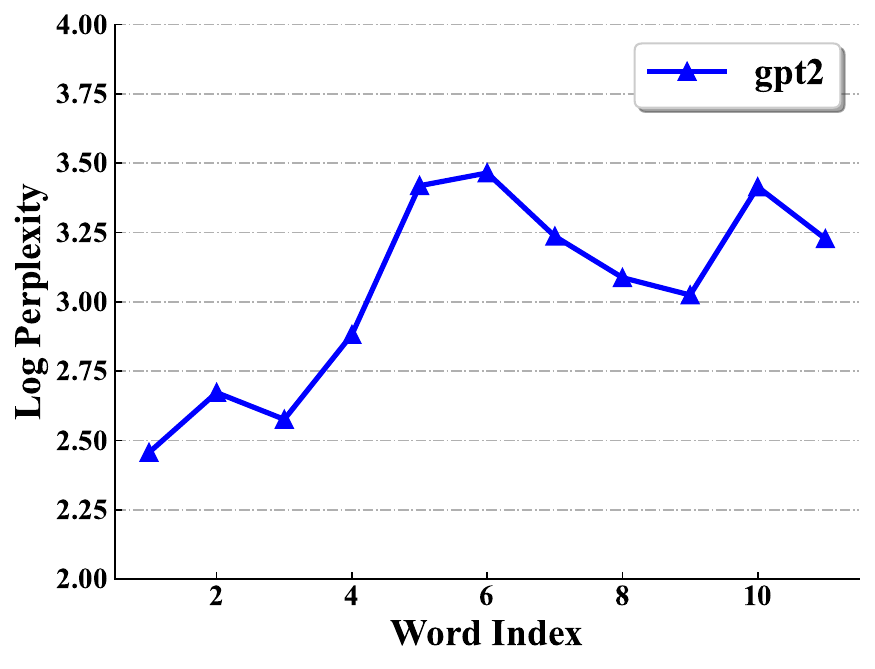}
\caption{Perplexity curve of gpt2 on the text "Turkish (Türkçe) is a language officially spoken in Turkey and Northern Cyprus." Each point represents the perplexity of the sequence from word index 0 to its respective word index.}
\label{figure1}
\end{figure}

Assessing the similarity of LLMs is a complex task, particularly when models are not fully open-source, a scenario that remains prevalent. In many cases, models may have publicly available parameter weights, yet their training datasets and processes are undisclosed. Although parameter space comparisons are feasible in such cases, estimating domain-specific model differences based solely on parameter comparisons remains a challenge. For LLMs with undisclosed parameters, existing methods evaluating similarity are underexplored. An intuitive approach involves comparing model outputs given identical prompts. However, even when outputs differ, models may still be similar due to underlying probability distributions. Another approach is to use evaluation benchmarks and metrics such as accuracy and BERTScore \cite{zhang2019bertscore} to assess performance similarity between models, but this method also lacks robustness, as shown in Table~\ref{table0}. If additional information, such as next-token probabilities, is available (e.g. the GPT-4 model \cite{achiam2023gpt}), alternative approaches can be applied. Since most auto-regressive LLMs are trained to maximize the likelihood of each token based on preceding tokens \cite{floridi2020gpt}, comparing the closeness of next-token distributions could provide insights into model similarity. However, the computational costs and applicability of such methods remain insufficiently studied.

\begin{table}[t]
\centering
\fontsize{10pt}{18pt}\selectfont
\resizebox{\columnwidth}{!}{%
\begin{tabular}{lllll}
\Xhline{1.2pt}
\textbf{Models} & \textbf{Llama(7b)} & \textbf{Pythia(6.9b)} & \textbf{Pythia(160m)} & \textbf{neo(125m)} \\ \hline
\textbf{PIQA}   & 0.798              & 0.76                  & 0.618                 & 0.631              \\ \Xhline{1.2pt}
\end{tabular}%
}
\caption{Zero-shot performance on the PIQA benchmark \cite{bisk2020piqa}. The scores of Llama7B and Pythia-6.9b on the PIQA benchmark are very close, as well as the scores of Pythia-160m and neo-125m, yet they are distinct LLMs.}
\label{table0}
\end{table}

Therefore, this work focuses on the similarity comparison of LLMs and its application across different domains, as well as its potential use in detecting model replication. We propose a novel metric to quantify LLM similarity, utilizing the perplexity curve (shown in Figure~\ref{figure1}) and the difference of Menger curvature \cite{leger1999menger} to represent the degree of similarity between LLMs.

To sum up, we make the following contributions in this work:
\begin{enumerate}
    \item We address the challenge of distinguishing a model coming from existing models by some simple methods, such as modifying model parameters or reformatting code. To this end, we develop a quantitative approach to measure LLM similarity.
    \item We validate the feasibility of our proposed metric through preliminary experiments and demonstrate that it outperforms several baseline methods. Furthermore, we expand our approach to include a broader range of LLMs and evaluation datasets across various domains. 
    \item We simulate a model-copying scenario by introducing noise into model parameters and establish thresholds for identifying copied LLMs, demonstrating the practical applicability of our method in real-world contexts.
\end{enumerate}

\section{Related work}
\label{sec2}
\subsection{Perplexity}
\label{subsec1}
Perplexity is a metric to measure the degree uncertainty in predicting the next token in a sequence based on preceding tokens. It is calculated using the negative average log-likelihood of texts under a language model \cite{brown1992estimate}. The formula for perplexity is defined as:
\begin{equation*}
    \text{PPL}(x)=exp[-\frac{1}{t}\sum_{i=1}^{t}\log p(x_i|x_{<i})]
\end{equation*}
where $x$ is a sequence of $t$ tokens, as described by \citet{alon2023detecting}. Here, $p(x_i|x_{<i})$ denotes the surprise of prediction, referring to the next-token probability discussed in Section~\ref{sec1}. Perplexity is widely used to detect LLM-generated texts \cite{tang2024science}. Research indicates that language models often concentrate on typical patterns in their training data, resulting in low perplexity scores for LLM-generated texts. In contrast, human-generated texts tend to exhibit higher perplexity values due to their varied styles of expression. Based on this observation, \citet{mitchell2023detectgpt} developed DetectGPT, employing probability curvature to detect machine-generated texts. DetectGPT's team finds that the change of log perplexity when applying perturbation to a text fragment is different for human-written texts and AI-generated texts. Building on the foundation of DetectGPT, \citet{xu2024detecting} designed AIGCode detector, which examines the perplexity change of code pieces after perturbation to discover AI-generated codes. While perplexity displays broad utility, research on its variation within a single sentence remains limited. Our method studies perplexity changes in a different manner and within a distinct application context.

\subsection{Pairwise Comparison of LLMs}
\label{subsec2}
While the evaluation of a single LLM is well-established \cite{liang2022holistic, chang2024survey}, recent research has begun to emphasize pairwise evaluations of LLMs. Motivated by the fact that comparing two options rather than scoring each one independently is more intuitive from a human perspective, \citet{liusie2023zero} examines comparative assessment across multiple dimensions, concluding that it offers a simple, general and effective approach for NLG (Natural Language Generation) assessment. \citet{kahng2024llm} introduces LLM Comparator, an innovative visual analytics tool for interactively analyzing results from automatic side-by-side evaluation, enabling detailed inspection of comparison details between two models. Despite the shift in focus from single model evaluation to multi-model evaluation, the study of LLM similarity remains an under-explored area.

\subsection{Data Privacy and Copyright in LLMs}
As the training corpus for LLMs continues to expand, studies increasingly focus on data privacy and copyright issues. Notable cases of privacy and copyright violations include Data Contamination \cite{sainz2023nlp}, also known as Benchmark Leakage \cite{zhou2023don}, and the illegal use of copyrighted and unauthorized data in training datasets. Data Contamination occurs when LLMs are trained on test data to artificially boost their scores and performance on evaluation metrics. Meanwhile, the presence of private and copyrighted materials in the training corpora of LLMs has sparked legal disputes, such as the lawsuit between \textit{The New York Times} and OpenAI \cite{nyt2023}, along with other cases \cite{Getty2023,Silverman2023}.

To address these concerns, methods such as Membership Inference (MI) \cite{shokri2017membership} and Dataset Inference (DI) \cite{maini2021dataset} have been developed. These techniques help determine if a particular dataset (DI) or data point (MI) is present in the training corpora, which can identify illegal dataset usage \cite{maini2024llm, shafran2021membership} and mitigate data contamination \cite{oren2023proving, shi2023detecting}. While previous research has primarily focused on ethical issues related to datasets, our work also considers model structures to uncover unethical practices and seek solutions to related problems.

\section{Approach}
\label{sec3}
Motivated by the observation that the perplexity of text segments may exhibit specific patterns after perturbations \cite{mitchell2023detectgpt}, we focus on analyzing the change in perplexity of a sentence segment when a small number of words are added or deleted. Given a word sequence consisting of $n$ words, denoted as $W_n = \{w_1, ..., w_n\}$, we compute the perplexity change around each word. Let $z$ be a symmetric integer random variable with $\mathbb{E}[z] = 0$. Based on the definition of perplexity, we define the perplexity change as follows:
\begin{equation}
\label{eq1}
\Delta \text{PPL}(w_i) = \log \text{PPL}(W_i) - \mathbb{E}_{z}[\log \text{PPL}(W_{i+z})]    
\end{equation}
Here, $W_i$ and $W_{i+z}$ denote the word sequences containing the first $i$ words and $i+z$ words, respectively. Let $\text{PPL}^A(\cdot)$ represent the perplexity calculated using model $A$, and $\text{PPL}^B(\cdot)$ represent the perplexity calculated using model $B$. We define the difference in perplexity change between models $A$ and $B$ on the sequence $W_n$ as the similarity value between the two models:
\begin{small}
\begin{equation}
\label{eq2}
\text{sim}(A,B,W_n)\!=\![\sum_{i=1}^{n} (\Delta \text{PPL}^A(w_i)-\Delta \text{PPL}^B(w_i))^2]^{\frac{1}{2}}
\end{equation}
\end{small}
However, directly calculating $\text{sim}(A, B, W_n)$ is difficult due to the unknown ground truth distribution of $z$. To address this, we approximate the distribution of $z$ by sampling from a simpler distribution. Let $x \in \{1, ..., n\}$ denote a word index, and define the function $f(x)$ as follows:
\begin{equation}
\label{eq3}
f(x) = \log \text{PPL}(W_{x})   
\end{equation}
Substituting the definition in Equation~\ref{eq3} into Equation~\ref{eq1}, we obtain:
\begin{equation*}
\Delta \text{PPL}(w_x) = f(x) - \mathbb{E}_{z} [f(x+z)]
\end{equation*}
Since $z$ is symmetric, we have $\mathbb{E}_{z} [f(x+z)] = \frac{1}{2} \mathbb{E}_{z} [f(x+z)+f(x-z)]$. Therefore, 
\begin{equation}
\label{eq4}
\Delta \text{PPL}(w_x) =  \mathbb{E}_{z} [f(x) - \frac{1}{2}(f(x+z)+f(x-z))]
\end{equation}
As $z$ measures the number of neighboring words considered when calculating the perplexity change, without loss of generality, we sample $\tilde{z}$ from a discrete uniform distribution $\tilde{z} \sim \text{Unif}\{-k,k\}$ and assign $z=\tilde{z}$, where $k$ is a positive integer close to 0. We obtain:
\begin{equation}
\label{eq5}
\Delta \text{PPL}(w_x) =  f(x) - \frac{1}{2}(f(x+\tilde{z})+f(x-\tilde{z}))
\end{equation}

\subsection*{Relation Between $\mathbf{\text{sim}(A,B,W_n)}$ and Menger Curvature Difference}
Let $\kappa(a, b, c)$ denote the Menger curvature of three points on the function $f(x)$ with $x$-coordinates $a$, $b$, and $c$. Similarly, let $A(a, b, c)$ represent the area of the triangle formed by these points, and let $l_{i,j}$ denote the chord length between two points on $f(x)$ with $x$-coordinates $i$ and $j$. By the definition of Menger curvature \cite{leger1999menger}, we have:
\begin{equation}
\label{eq6}
\kappa(x-\tilde{z},x,x+\tilde{z}) = \frac{4A(x-\tilde{z},x,x+\tilde{z})}{l_{x-\tilde{z},x}l_{x,x+\tilde{z}}l_{x-\tilde{z},x+\tilde{z}}}
\end{equation}
Using the determinant formula to calculate the area of a triangle, we derive:
\begin{small}
\begin{equation}
\label{eq7}
A(x-\tilde{z},x,x+\tilde{z}) = |\frac{\tilde{z}}{2}[f(x+\tilde{z})+f(x-\tilde{z})-2f(x)]|
\end{equation}
\end{small}
Here, $|\cdot|$ denotes the absolute value. Combining Equation~\ref{eq5} and Equation~\ref{eq7}, we obtain:
\begin{equation}
\label{eq8}
A(x-\tilde{z},x,x+\tilde{z}) = |\tilde{z}\Delta \text{PPL}(w_x)|
\end{equation}
Since the Menger curvature of any triple of points is always positive, we introduce an indicator function to capture the sign of $\Delta \text{PPL}(w_x)$:
\begin{equation*}
\mathbb{I}(f,x,y) = \text{sgn}(f(x) - \frac{1}{2}[f(x+y) + f(x-y)])
\end{equation*}
where the sign function $\text{sgn}(u)$ is defined as:
\begin{small}
\begin{equation*}
\text{sgn}(u) = 
\begin{cases} 
1 & \text{if } u \geq 0, \\
-1 & \text{if } u < 0.
\end{cases}
\end{equation*}
\end{small}
Applying the indicator function and substituting Equation~\ref{eq8} into the Menger curvature definition in Equation~\ref{eq6}, we obtain:
\begin{small}
\begin{align}
\label{eq9}
&|\Delta \text{PPL}^A(w_x)-\Delta \text{PPL}^B(w_x)| = \nonumber\\ 
&\frac{1}{4\tilde{z}}|l^A_{x-\tilde{z},x}l^A_{x,x+\tilde{z}}l^A_{x-\tilde{z},x+\tilde{z}}\mathbb{I}(f_A,x,\tilde{z})\kappa^A(x-\tilde{z},x,x+\tilde{z})\nonumber \\
&-l^B_{x-\tilde{z},x}l^B_{x,x+\tilde{z}}l^B_{x-\tilde{z},x+\tilde{z}}\mathbb{I}(f_B,x,\tilde{z})\kappa^B(x-\tilde{z},x,x+\tilde{z})|
\end{align}
\end{small}
Given that $f(x)$ is discrete and finite, we can identify an upper bound $U$ such that:
\begin{equation*} 
l^A_{x-\tilde{z}, x} l^A_{x, x+\tilde{z}} l^A_{x-\tilde{z}, x+\tilde{z}} \leq U, \quad \forall x 
\end{equation*}
Furthermore, we make the following assumption:
\begin{equation}
\label{eq10}
\frac{l^B_{x-\tilde{z},x}l^B_{x,x+\tilde{z}}l^B_{x-\tilde{z},x+\tilde{z}}}{l^A_{x-\tilde{z},x}l^A_{x,x+\tilde{z}}l^A_{x-\tilde{z},x+\tilde{z}}} \approx 1, \quad \forall x
\end{equation}
Finally, we derive the following upper bound for the similarity formula:
\begin{small}
\begin{align}
\label{eq11}
&\text{sim}(A,B,W_n)\leq\frac{U}{4\tilde{z}}[\sum_{i=1}^{n} (\mathbb{I}(f_A,i,\tilde{z})\kappa^A(i-\tilde{z},i,i+\tilde{z})\nonumber\\
&- \mathbb{I}(f_B,i,\tilde{z})\kappa^B(i-\tilde{z},i,i+\tilde{z}))^2]^{\frac{1}{2}}
\end{align}
\end{small}

\begin{figure}[t]
\centering
\includegraphics[width=1.0\columnwidth]{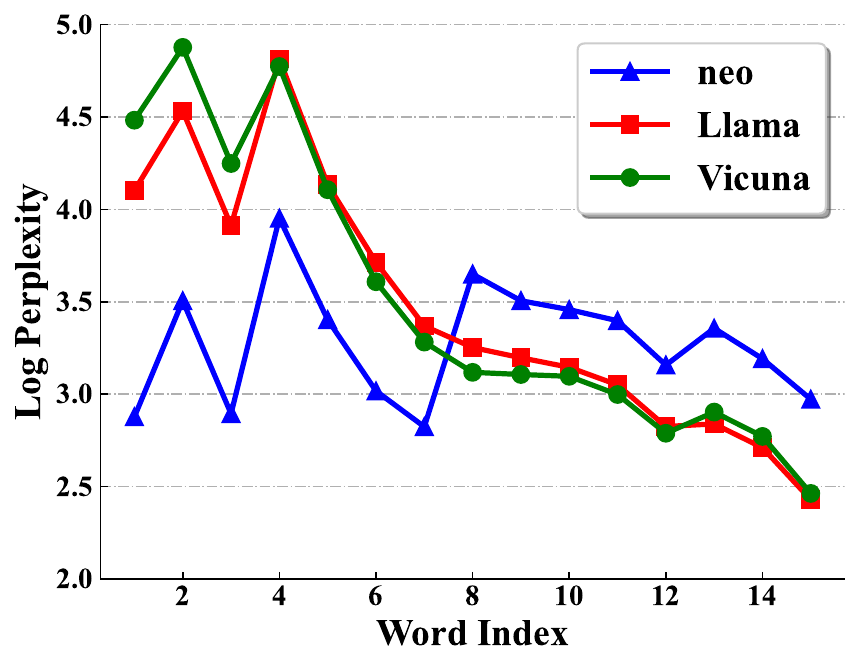}
\caption{Perplexity curves of Llama7B, Vicuna7B, and gpt-neo-125M on the text "Associations between age and gray matter volume in anatomical brain networks in middle-aged to older adults."}
\label{figure2}
\end{figure}

Equation~\ref{eq11} demonstrates that the similarity value between two LLMs can be upper bounded by their Menger curvature difference. A smaller upper bound indicates a lower similarity value, suggesting that the two models are more closely related. For instance, in the case of Llama, Vicuna, and neo, since Vicuna is fine-tuned on Llama, it is expected to be more similar to Llama than to neo. As shown in Figure~\ref{figure2}, the perplexity curves for Llama and Vicuna have more comparable Menger curvature, indicating a lower similarity value and a closer relationship between the two models.

Moreover, the validity of the assumption in Equation~\ref{eq10} can be approximately verified from Figure~\ref{figure2}, as the product of chord lengths between neighboring points doesn't differ too much for all word indices across LLMs. Alternatively, the similarity value between two LLMs can be computed directly using Equation~\ref{eq5}. We provide a detailed comparison of different similarity computation methods in Section~\ref{sec4.3}.

\begin{table*}[htbp]
\centering
\resizebox{\linewidth}{!}{ 
\fontsize{10pt}{18pt}\selectfont
\begin{tabular}{cccc}
\Xhline{1.2pt}
\textbf{Models / Similarity} & \textbf{gpt2 (openwebtext[:1M])} & \textbf{gpt2 (openwebtext[1M:2M])} & \textbf{gpt2 (pile[:1M])} \\ \hline
\textbf{gpt2 (openwebtext[:1M])}     & /      & 0.3579 & 0.6895 \\ 
\textbf{gpt2 (openwebtext[1M:2M])}   & 0.3579 & /      & 0.6932 \\ 
\textbf{gpt2 (pile[:1M])}            & 0.6895 & 0.6932 & /      \\ 
\Xhline{1.2pt}
\end{tabular}
}
\caption{Similarity of \textbf{gpt2-124m} trained on datasets from different distributions.}
\label{table2}
\end{table*}

\begin{table*}[htbp]
\centering
\resizebox{\linewidth}{!}{ 
\fontsize{10pt}{18pt}\selectfont
\begin{tabular}{cccc}
\Xhline{1.2pt}
\textbf{Models / Similarity} & \textbf{Pythia (openwebtext[:1M])} & \textbf{Pythia (openwebtext[1M:2M])} & \textbf{Pythia (pile[:1M])} \\ \hline
\textbf{Pythia (openwebtext[:1M])}     & /      & 0.3823 & 0.6243 \\ 
\textbf{Pythia (openwebtext[1M:2M])} & 0.3823  & /     & 0.6276  \\ 
\textbf{Pythia (pile[:1M])} & 0.6243 & 0.6276 & /      \\ \Xhline{1.2pt}
\end{tabular}
}
\caption{Similarity of \textbf{Pythia-70m} trained on datasets from different distributions.}
\label{table3}
\end{table*}

\begin{table}[htbp]
\centering
\resizebox{\columnwidth}{!}{ 
\fontsize{10pt}{18pt}\selectfont
\begin{tabular}{ccc}
\Xhline{1.2pt}
\textbf{Models / Similarity} & \textbf{gpt2-124m-modified} & \textbf{opt-125m} \\ \hline
\textbf{gpt2-124m}                                          & 0.3156                        & 0.4648                       \\ 
\textbf{opt-125m-modified}                                         & 0.4485                        & 0.3055                       \\ \Xhline{1.2pt}
\end{tabular}
}
\caption{Similarity of LLMs trained on the first 1M samples of the OpenWebText corpus.}
\label{table4}
\end{table}

\section{Experiments}
\label{sec4}
The experiments select 1000 samples in each run of similarity computation. We adopt $k=1$ in the discrete uniform distribution and sample $\tilde{z}$ once for each word index. Denote the set formed by these 1000 samples as $\Omega$. The similarity between Model A and Model B is given by:
\begin{equation*}
\text{sim}(A,B) = \frac{\sum\limits_{W_n \in \Omega} |W_n| \cdot \text{sim}(A,B,W_n)}{\sum\limits_{W_n \in \Omega} |W_n|}   \\
\end{equation*}
We use a weighted average of the similarity values for each sample, with the weights based on the length of the sample, where $|W_n|$ denotes the cardinality of the word sequence. By applying the inequality in Equation~\ref{eq11}, the overall similarity between models $A$ and $B$, denoted as $\text{sim}(A, B)$, can be upper bounded by the difference in their Menger curvatures.

\subsection{Preliminary Experiments}
\label{subsec4}
We first use preliminary experiments to demonstrate the feasibility of our approach. Since most LLMs are trained on a wide range of datasets and the details of their pre-training and fine-tuning are often not publicly available, it is challenging to determine whether our approach can accurately reflect similarity of the models. Therefore, we train some small-sized LLMs from scratch and use our approach to analyze their similarity. We perform similarity calculations on the Wikipedia dataset\footnote{\href{https://huggingface.co/datasets/legacy-datasets/wikipedia}{legacy-datasets/wikipedia}}, as the training datasets also contain general knowledge. The differences among LLMs can generally be divided into two categories. 
\begin{itemize}
    \item Suppose the model size doesn't vary much, if we fix the training dataset, LLMs of the same model suite and slightly different sizes will be more similar than LLMs of different model suites but the same size.
    \item If we fix the size and model suite, LLMs trained on in-distribution datasets will be more similar than LLMs trained on out-of-distribution datasets.
\end{itemize}
Based on these two categories, we design two scenarios and test our approach on each one.

\begin{table*}[htbp]
\centering
\setlength{\tabcolsep}{6pt}
\renewcommand{\arraystretch}{1.2}
\resizebox{\linewidth}{!}{
\fontsize{8pt}{10pt}\selectfont
\begin{tabular}{ccccc|ccccc}
\Xhline{1.2pt}
\textbf{Model1} &
  \textbf{Model2} &
  \textbf{JSD} &
  \textbf{Sim\_Approx} &
  \textbf{Ours} &
  \textbf{Model1} &
  \textbf{Model2} &
  \textbf{JSD} &
  \textbf{Sim\_Approx} &
  \textbf{Ours} \\ \hline
\textbf{gpt2}        & \textbf{gpt2(medium)} & {\ul 0.092}  & {\ul 0.4288} & {\ul 0.4567} & \textbf{opt(125m)}    & \textbf{Pythia(160m)} & 0.1376 & 0.7224 & 0.6228 \\
\textbf{gpt2}        & \textbf{opt(125m)}    & 0.1456 & 0.6215 & 0.628  & \textbf{opt(125m)}    & \textbf{Pythia(6.9b)} & 0.1975 & 0.7487 & 0.7931 \\
\textbf{gpt2}        & \textbf{neo(125m)}    & 0.1386 & {\ul 0.5412} & 0.5916 & \textbf{opt(125m)}    & \textbf{Dolly(v2,7b)} & 0.264  & 0.8832 & 0.9228 \\
\textbf{gpt2}        & \textbf{Pythia(160m)} & 0.1374 & 0.7275 & 0.6663 & \textbf{opt(125m)}    & \textbf{Dolly(v1,6b)} & 0.2153 & 0.7484 & 0.8013 \\
\textbf{gpt2}        & \textbf{Pythia(6.9b)} & 0.2491 & 0.8222 & 0.9394 & \textbf{neo(125m)}    & \textbf{Pythia(160m)} & {\ul 0.1235} & 0.6535 & {\ul 0.5763} \\
\textbf{gpt2}        & \textbf{Dolly(v2,7b)} & 0.3228 & 0.9686 & 1.0695 & \textbf{neo(125m)}    & \textbf{Pythia(6.9b)} & 0.1979 & 0.7501 & 0.8147 \\
\textbf{gpt2}        & \textbf{Dolly(v1,6b)} & 0.2728 & 0.8044 & 0.9282 & \textbf{neo(125m)}    & \textbf{Dolly(v2,7b)} & 0.2671 & 0.8838 & 0.9432 \\
\textbf{gpt2(medium)} & \textbf{opt(125m)}    & 0.1357 & 0.6471 & 0.6485 & \textbf{neo(125m)}    & \textbf{Dolly(v1,6b)} & 0.211  & 0.6977 & 0.7927 \\
\textbf{gpt2(medium)} & \textbf{neo(125m)}    & 0.1344 & {\ul 0.5578} & 0.6177 & \textbf{Pythia(160m)} & \textbf{Pythia(6.9b)} & 0.1928 & 0.716  & 0.7394 \\
\textbf{gpt2(medium)} & \textbf{Pythia(160m)} & 0.1425 & 0.7435 & 0.6892 & \textbf{Pythia(160m)} & \textbf{Dolly(v2,7b)} & 0.2718 & 0.8902 & 0.9084 \\
\textbf{gpt2(medium)} & \textbf{Pythia(6.9b)} & 0.1957 & 0.7416 & 0.8393 & \textbf{Pythia(160m)} & \textbf{Dolly(v1,6b)} & 0.2457 & 0.8871 & 0.8578 \\
\textbf{gpt2(medium)} & \textbf{Dolly(v2,7b)} & 0.268  & 0.8892 & 0.9795 & \textbf{Pythia(6.9b)} & \textbf{Dolly(v2,7b)} & {\ul \textbf{0.0714}} & {\ul \textbf{0.3798}} & {\ul \textbf{0.4114}} \\
\textbf{gpt2(medium)} & \textbf{Dolly(v1,6b)} & 0.2128 & 0.7126 & 0.8148 & \textbf{Pythia(6.9b)} & \textbf{Dolly(v1,6b)} & {\ul 0.1039} & {\ul 0.5408} & {\ul 0.5142} \\
\textbf{opt(125m)}    & \textbf{neo(125m)}    & {\ul 0.1186} & 0.5664 & {\ul 0.5429} & \textbf{Dolly(v2,7b)} & \textbf{Dolly(v1,6b)} & 0.1433 & 0.6623 & 0.6378 \\ \Xhline{1.2pt}
\end{tabular}
}
\caption{Evaluation of baseline methods and our approach. The smallest similarity value of each method is put in bold, and the least five are underlined.}
\label{baseline}
\end{table*}

\subsubsection{Scenario One: Adjusting the Model Structure}
\label{subsub1}
In this section, we fix the training dataset and vary the model structures to assess whether our approach can reflect the degree of structural changes in models. Specifically, We use the first 1M samples from the OpenWebText corpus \cite{Gokaslan2019OpenWeb}
as the training dataset and select two base models: gpt2-124m \cite{radford2019language,brown2020language} and opt-125m \cite{zhang2022opt}. Four LLMs are trained from scratch: \textbf{gpt2-124m}, \textbf{gpt2-124m-modified}, \textbf{opt-125m}, and \textbf{opt-125m-modified}. The modified versions of both models reduce the number of hidden layers in the Transformer encoder by one, while keeping other structural components unchanged. Given that this modification is minimal—resulting in a parameter reduction of less than 5M—the gpt2-124m model is expected to be more similar to its modified counterpart than to the opt models. Using Menger curvature differences to represent model similarity, we obtain the results shown in Table~\ref{table4}.

From Table~\ref{table4}, we can observe that models within the same family have smaller similarity values compared to models from different families, which aligns with our expectations. Therefore, our approach can reflect the differences in LLMs caused by variations in model structures.

\subsubsection{Scenario Two: Adjusting the Distribution of Training Datasets}
\label{subsub2}
\paragraph{Domain Generalization and Distribution Shift}
\label{para1}
Every dataset is sampled from a data-generating distribution (an unknown distribution under a data-generating process). Real-life documents in different areas, such as news and novels, often exhibit distinct characteristics and originate from different data-generating distributions \cite{hendrycks2020pretrained}. This phenomenon, known as natural distribution shift, is closely related to domain generalization \cite{li2023survey}. Recent studies highlight that LLMs often struggle with domain generalization, showing limited abilities to generalize beyond in-distribution test data and frequently performing poorly on out-of-distribution data \cite{ebrahimi2017hotflip,hendrycks2020pretrained,gururangan2018annotation}. Given these findings, it is expected that LLMs trained on datasets from the same distribution will exhibit greater similarity than those trained on datasets from different data-generating distributions.

Therefore, in this section, we change the data-generating distribution of training datasets to assess whether our method can detect distribution shifts. We utilize the Pile \cite{gao2020pile}, a general-purpose dataset containing texts from 22 diverse sources. Similar to OpenWebText, the dataset also involves general knowledge, but originates from a different data-generating distribution. Given the large size of the Pile dataset (approximately 800GB), we limit our usage to the first 1M samples. Additionally, we create two subsets from the OpenWebText corpus: the first 1M samples and the 1M to 2M samples. Both subsets are drawn from the same underlying data-generating distribution. In total, three training datasets are used: one subset from the Pile and two subsets from OpenWebText. These datasets are employed to train the gpt2-124m model and the Pythia-70m model \cite{biderman2023Pythia}.

Tables~\ref{table2} and \ref{table3} present the similarity evaluations of the trained models, with the brackets indicating the specific dataset subsets used for training. The results show that the similarity scores between LLMs trained on the two subsets of OpenWebText are notably lower compared to the scores between LLMs trained on one subset of OpenWebText and one subset of the Pile. This highlights the effectiveness of our approach in capturing differences in LLMs caused by distribution shifts in training datasets.

\begin{table*}[htbp]
\centering
\setlength{\tabcolsep}{6pt}
\renewcommand{\arraystretch}{1.2}
\resizebox{\linewidth}{!}{
\fontsize{8pt}{10pt}\selectfont
\begin{tabular}{cccccccc}
\Xhline{1.2pt}
 \begin{tabular}[c]{@{}c@{}}\vspace{0pt}\\ \vspace{0pt} \end{tabular}&
  \textbf{\begin{tabular}[c]{@{}c@{}}Pythia(160m)\\ [-3pt]\& gpt2\end{tabular}} &
  \textbf{\begin{tabular}[c]{@{}c@{}}Pythia(160m) \&\\ [-3pt] gpt2(medium)\end{tabular}} &
  \textbf{\begin{tabular}[c]{@{}c@{}}Pythia(160m)\\ [-3pt] \& opt(125m)\end{tabular}} &
  \textbf{\begin{tabular}[c]{@{}c@{}}Pythia(160m)\\ [-3pt] \& neo(125m)\end{tabular}} &
  \textbf{\begin{tabular}[c]{@{}c@{}}Pythia(160m) \&\\ [-3pt] Pythia(6.9b)\end{tabular}} &
  \textbf{\begin{tabular}[c]{@{}c@{}}Pythia(160m) \&\\ [-3pt] Dolly(v2,7b)\end{tabular}} &
  \textbf{\begin{tabular}[c]{@{}c@{}}Pythia(160m) \&\\ [-3pt] Dolly(v1,6b)\end{tabular}} \\ \hline
\textbf{JSD}         & 0.0049 & 0.0044 & 0.0042 & 0.0037 & 0.0063 & 0.0079 & 0.0071 \\
\textbf{Sim\_Approx} & 0.1309 & 0.1303 & 0.1293 & 0.1305 & 0.125  & 0.1269 & 0.1318 \\
\textbf{Ours}        & 0.0358 & 0.0333 & 0.0314 & 0.0302 & 0.0364 & 0.0406 & 0.0406 \\ \Xhline{1.2pt}
\end{tabular}
}
\caption{Standard deviation of different methods and LLM pairs across the selected samples in each run (measured on the Wikipedia dataset).}
\label{variance}
\end{table*}

\begin{table*}[htbp]
\centering
\renewcommand{\arraystretch}{1.2}
\fontsize{8pt}{10pt}\selectfont
\begin{tabular}{cccc}
\Xhline{1.2pt}
 &
  \textbf{JSD} &
  \textbf{Sim\_Approx} &
  \textbf{Ours} \\ \hline
\textbf{Model Info Requirement} &
  \begin{tabular}[c]{@{}c@{}}Require distribution of\\ [-2pt] all tokens\end{tabular} &
  \begin{tabular}[c]{@{}c@{}}Require perplexity\\ [-2pt] (Highest token probability)\end{tabular} &
  \begin{tabular}[c]{@{}c@{}}Require perplexity\\ [-2pt] (Highest token probability)\end{tabular} \\
\textbf{\begin{tabular}[c]{@{}c@{}}Variation across\\ [-2pt] Different Samples\end{tabular}} &
  Low &
  Relatively high &
  Relatively low \\
\textbf{Range of Application} &
  \begin{tabular}[c]{@{}c@{}}LLM pairs with high\\ [-2pt] vocabulary overlap\end{tabular} &
  All LLM pairs &
  All LLM pairs \\
\textbf{\begin{tabular}[c]{@{}c@{}}Computational Complexity on a\\ [-2pt] Word Sequence with Length N\end{tabular}} &
  $\mathcal{O}(N\Bar{V})$ &
  $\mathcal{O}(N)$ &
  $\mathcal{O}(N)$ \\ \Xhline{1.2pt}
\end{tabular}
\caption{Comparison of baseline methods and our approach. To estimate computational complexity, the \textbf{basic operation} is defined as a constant-time arithmetic or geometric calculation performed on pairs or triplets of consecutive points, including distance and area calculations. $\Bar{V}$ denotes the overlapped vocabulary size of LLM pairs.}
\label{comparison}
\end{table*}

\subsection{Baseline Experiments}
\label{sec4.3}
After confirming feasibility, we use baseline experiments to demonstrate the superiority of our method. Since LLM similarity evaluation is not well-established, we consider two intuitive baselines for this task.
\subsubsection{Similarity Approximation}
 The first baseline is Similarity Approximation (\textbf{Sim\_Approx}), which leverages Equation~\ref{eq5} described in Section~\ref{sec3}. Applying the discrete uniform distribution, we can approximate the ground truth distribution of $z$ which may be complicated in real cases. The key distinction between Similarity Approximation and our approach lies in the use of Menger Curvature, as both methods require the derivation of perplexity curves. We use $k=1$ in the discrete uniform distribution and sample $\tilde{z}$ once for each word index to ensure consistency with our approach.
\subsubsection{Jensen-Shannon Divergence}
The second baseline is the Jensen-Shannon Divergence (\textbf{JSD}) \cite{menendez1997jensen}, which measures the divergence between next-token probability distributions of LLMs. Unlike perplexity, which considers the probability of observed tokens in a word sequence, next-token distributions account for the probability of all tokens in the vocabulary, providing a broader perspective on a model's characteristics. Consequently, another intuitive way to compare two LLMs' similarity is to evaluate the closeness of their next-token distributions across text sequences. However, a challenge arises when comparing models with different vocabularies, as their next-token distributions involve inconsistent numbers of random variables. To solve this problem, we calculate the vocabulary overlap between two LLMs and select pairs with an overlap greater than $70\%$, i.e.,
\begin{equation*}
\frac{2*\text{n\_overlapped\_vocab}}{\text{n\_LLM1\_vocab} + \text{n\_LLM2\_vocab}} \geq 0.7 
\end{equation*}
For symmetry, we use Jensen-Shannon Divergence to measure the distribution similarity. Given a text sequence, we extract all sub-sequences by word, calculate the JSD for next-token distributions of the last token in each sub-sequence, and average these values to represent the models' similarity on the text sequence.

\begin{table*}[htbp]
\centering
\fontsize{8pt}{15pt}\selectfont
\resizebox{\linewidth}{!}{%
\begin{tabular}{cccccccccc}
\Xhline{1.2pt}
\textbf{\fontsize{8pt}{10pt}\selectfont\begin{tabular}[c]{@{}c@{}}Models/\\  Similarity\end{tabular}} &
  \textbf{gpt2} &
  \textbf{\fontsize{8pt}{10pt}\selectfont\begin{tabular}[c]{@{}c@{}}gpt2\\ [-2pt](medium)\end{tabular}} &
  \textbf{\fontsize{8pt}{10pt}\selectfont\begin{tabular}[c]{@{}c@{}}Llama\\ [-2pt](7b)\end{tabular}} &
  \textbf{\fontsize{8pt}{10pt}\selectfont\begin{tabular}[c]{@{}c@{}}neo\\ [-2pt](125m)\end{tabular}} &
  \textbf{\fontsize{8pt}{10pt}\selectfont\begin{tabular}[c]{@{}c@{}}Vicuna\\ [-2pt](7b)\end{tabular}} &
  \textbf{\fontsize{8pt}{10pt}\selectfont\begin{tabular}[c]{@{}c@{}}Pythia\\ [-2pt](160m)\end{tabular}} &
  \textbf{\fontsize{8pt}{10pt}\selectfont\begin{tabular}[c]{@{}c@{}}Pythia\\ [-2pt](6.9b)\end{tabular}} &
  \textbf{\fontsize{8pt}{10pt}\selectfont\begin{tabular}[c]{@{}c@{}}Dolly\\ [-2pt](v2,7b)\end{tabular}} &
  \textbf{\fontsize{8pt}{10pt}\selectfont\begin{tabular}[c]{@{}c@{}}Dolly\\ [-2pt](v1,6b)\end{tabular}} \\ \hline
\textbf{gpt2}         & /           & {\ul 0.4567} & 1.2336       & 0.5916      & 1.2841       & 0.6663      & 0.9394                & 1.0695                 & 0.9282       \\ 
\textbf{gpt2(medium)} & {\ul 0.4567} & /           & 1.2263       & 0.6177      & 1.2332       & 0.6892       & 0.8393                & 0.9795                & 0.8148       \\ 
\textbf{Llama(7b)}    & 1.2336      & 1.2263      & /            & 1.1922      & {\ul 0.4652} & 1.1548      & 0.9658                & 1.0359                & 1.0133       \\ 
\textbf{neo(125m)}    & 0.5916      & 0.6177      & 1.1922       & /           & 1.2487       & {\ul 0.5763} & 0.8147                & 0.9432                & 0.7927       \\ 
\textbf{Vicuna(7b)}   & 1.2841      & 1.2332      & {\ul 0.4652} & 1.2487      & /            & 1.1716      & 0.9581                & 1.0799                & 0.9699       \\ 
\textbf{Pythia(160m)} & 0.6663      & 0.6892       & 1.1548       & {\ul 0.5763} & 1.1716       & /           & 0.7394                & 0.9084                 & 0.8578        \\ 
\textbf{Pythia(6.9b)} & 0.9394      & 0.8393      & 0.9658       & 0.8147      & 0.9581       & 0.7394      & /                     & {\ul \textbf{0.4114}} & {\ul 0.5142} \\ 
\textbf{Dolly(v2,7b)} & 1.0695       & 0.9795      & 1.0359       & 0.9432      & 1.0799       & 0.9084       & {\ul \textbf{0.4114}} & /                     & 0.6378       \\ 
\textbf{Dolly(v1,6b)} & 0.9282      & 0.8148      & 1.0133       & 0.7927      & 0.9699       & 0.8578       & {\ul 0.5142}          & 0.6378                & /            \\ \Xhline{1.2pt}
\end{tabular}%
}
\caption{LLM similarity on the Wikipedia dataset. The smallest similarity value is put in bold, the least five are underlined.}
\label{table5}
\end{table*}

\begin{table*}[t]
\centering
\fontsize{8pt}{15pt}\selectfont
\resizebox{\linewidth}{!}{%
\begin{tabular}{ccccccccccccccccccc}
\Xhline{1.2pt}
\multicolumn{1}{c}{\textbf{Models/}} &
  \multicolumn{9}{c}{\textbf{Med}} &
  \multicolumn{9}{c}{\textbf{Law}} \\ \cline{2-19}
\textbf{\fontsize{8pt}{10pt}\selectfont\begin{tabular}[c]{@{}c@{}}Similarity\\ \vspace{0pt} \end{tabular}} &
  \textbf{gpt2} &
  \textbf{\fontsize{8pt}{10pt}\selectfont\begin{tabular}[c]{@{}c@{}}gpt2\\ [-2pt](medium)\end{tabular}} &
  \textbf{\fontsize{8pt}{10pt}\selectfont\begin{tabular}[c]{@{}c@{}}Llama\\ [-2pt](7b)\end{tabular}} &
  \textbf{\fontsize{8pt}{10pt}\selectfont\begin{tabular}[c]{@{}c@{}}neo\\ [-2pt](125m)\end{tabular}} &
  \textbf{\fontsize{8pt}{10pt}\selectfont\begin{tabular}[c]{@{}c@{}}Vicuna\\ [-2pt](7b)\end{tabular}} &
  \textbf{\fontsize{8pt}{10pt}\selectfont\begin{tabular}[c]{@{}c@{}}Pythia\\ [-2pt](160m)\end{tabular}} &
  \textbf{\fontsize{8pt}{10pt}\selectfont\begin{tabular}[c]{@{}c@{}}Pythia\\ [-2pt](6.9b)\end{tabular}} &
  \textbf{\fontsize{8pt}{10pt}\selectfont\begin{tabular}[c]{@{}c@{}}Dolly\\ [-2pt](v2,7b)\end{tabular}} &
  \multicolumn{1}{c|}{\fontsize{8pt}{10pt}\selectfont\textbf{\begin{tabular}[c]{@{}c@{}}Dolly\\ [-2pt](v1,6b)\end{tabular}}} &
  \textbf{gpt2} &
  \textbf{\fontsize{8pt}{10pt}\selectfont\begin{tabular}[c]{@{}c@{}}gpt2\\ [-2pt](medium)\end{tabular}} &
  \textbf{\fontsize{8pt}{10pt}\selectfont\begin{tabular}[c]{@{}c@{}}Llama\\ [-2pt](7b)\end{tabular}} &
  \textbf{\fontsize{8pt}{10pt}\selectfont\begin{tabular}[c]{@{}c@{}}neo\\ [-2pt](125m)\end{tabular}} &
  \textbf{\fontsize{8pt}{10pt}\selectfont\begin{tabular}[c]{@{}c@{}}Vicuna\\ [-2pt](7b)\end{tabular}} &
  \textbf{\fontsize{8pt}{10pt}\selectfont\begin{tabular}[c]{@{}c@{}}Pythia\\ [-2pt](160m)\end{tabular}} &
  \textbf{\fontsize{8pt}{10pt}\selectfont\begin{tabular}[c]{@{}c@{}}Pythia\\ [-2pt](6.9b)\end{tabular}} &
  \textbf{\fontsize{8pt}{10pt}\selectfont\begin{tabular}[c]{@{}c@{}}Dolly\\ [-2pt](v2,7b)\end{tabular}} &
  \textbf{\fontsize{8pt}{10pt}\selectfont\begin{tabular}[c]{@{}c@{}}Dolly\\ [-2pt](v1,6b)\end{tabular}} \\ \hline
\textbf{gpt2} &
  / &
  {\ul 0.4365} &
  1.0452 &
  0.6014 &
  1.1257 &
  0.7981 &
  0.9013 &
  1.0111 &
  \multicolumn{1}{c|}{0.8415} &
  / &
  {\ul \textbf{0.5247}} &
  1.3733 &
  0.7201 &
  1.3713 &
  0.7951 &
  1.0433 &
  1.1560 &
  1.1924 \\ 
\textbf{\fontsize{8pt}{10pt}\selectfont\begin{tabular}[c]{@{}c@{}}gpt2\\ [-2pt](medium)\end{tabular}} &
  {\ul 0.4365} &
  / &
  0.9602 &
  {\ul 0.5255} &
  1.0391 &
  0.7513 &
  0.8167 &
  0.9418 &
  \multicolumn{1}{c|}{0.7181} &
  {\ul \textbf{0.5247}} &
  / &
  1.2643 &
  0.8176 &
  1.2862 &
  0.8323 &
  0.9694 &
  1.1138 &
  1.0775 \\ 
\textbf{\fontsize{8pt}{10pt}\selectfont\begin{tabular}[c]{@{}c@{}}Llama\\ [-2pt](7b)\end{tabular}} &
  1.0452 &
  0.9602 &
  / &
  0.9641 &
  {\ul 0.4405} &
  1.0435 &
  0.8232 &
  0.9409 &
  \multicolumn{1}{c|}{0.7764} &
  1.3733 &
  1.2643 &
  / &
  1.2571 &
  {\ul 0.5407} &
  1.1962 &
  0.9713 &
  1.1183 &
  1.0633 \\ 
\textbf{\fontsize{8pt}{10pt}\selectfont\begin{tabular}[c]{@{}c@{}}neo\\ [-2pt](125m)\end{tabular}} &
  0.6014 &
  {\ul 0.5255} &
  0.9641 &
  / &
  1.0317 &
  0.6696 &
  0.7372 &
  0.8463 &
  \multicolumn{1}{c|}{0.6654} &
  0.7201 &
  0.8176 &
  1.2571 &
  / &
  1.3783 &
  {\ul 0.6250} &
  0.9132 &
  1.0886 &
  0.9869 \\ 
\textbf{\fontsize{8pt}{10pt}\selectfont\begin{tabular}[c]{@{}c@{}}Vicuna\\ [-2pt](7b)\end{tabular}} &
  1.1257 &
  1.0391 &
  {\ul 0.4405} &
  1.0317 &
  / &
  1.1042 &
  0.8843 &
  0.9656 &
  \multicolumn{1}{c|}{0.8082} &
  1.3713 &
  1.2862 &
  {\ul 0.5407} &
  1.3783 &
  / &
  1.3207 &
  1.0221 &
  1.1450 &
  1.1424 \\ 
\textbf{\fontsize{8pt}{10pt}\selectfont\begin{tabular}[c]{@{}c@{}}Pythia\\ [-2pt](160m)\end{tabular}} &
  0.7981 &
  0.7513 &
  1.0435 &
  0.6696 &
  1.1042 &
  / &
  0.6296 &
  0.7547 &
  \multicolumn{1}{c|}{0.7972} &
  0.7951 &
  0.8323 &
  1.1962 &
  {\ul 0.6250} &
  1.3207 &
  / &
  0.8315 &
  1.0410 &
  1.0128 \\ 
\textbf{\fontsize{8pt}{10pt}\selectfont\begin{tabular}[c]{@{}c@{}}Pythia\\ [-2pt](6.9b)\end{tabular}} &
  0.9013 &
  0.8167 &
  0.8232 &
  0.7372 &
  0.8843 &
  0.6296 &
  / &
  {\ul \textbf{0.3446}} &
  \multicolumn{1}{c|}{{\ul 0.5181}} &
  1.0433 &
  0.9694 &
  0.9713 &
  0.9132 &
  1.0221 &
  0.8315 &
  / &
  {\ul 0.5522} &
  {\ul 0.6945} \\ 
\textbf{\fontsize{8pt}{10pt}\selectfont\begin{tabular}[c]{@{}c@{}}Dolly\\ [-2pt](v2,7b)\end{tabular}} &
  1.0111 &
  0.9418 &
  0.9409 &
  0.8463 &
  0.9656 &
  0.7547 &
  {\ul \textbf{0.3446}} &
  / &
  \multicolumn{1}{c|}{0.6268} &
  1.1560 &
  1.1138 &
  1.1183 &
  1.0886 &
  1.1450 &
  1.0410 &
  {\ul 0.5522} &
  / &
  0.8596 \\ 
\textbf{\fontsize{8pt}{10pt}\selectfont\begin{tabular}[c]{@{}c@{}}Dolly\\ [-2pt](v1,6b)\end{tabular}} &
  0.8415 &
  0.7181 &
  0.7764 &
  0.6654 &
  0.8082 &
  0.7972 &
  {\ul 0.5181} &
  0.6268 &
  \multicolumn{1}{c|}{/} &
  1.1924 &
  1.0775 &
  1.0633 &
  0.9869 &
  1.1424 &
  1.0128 &
  {\ul 0.6945} &
  0.8596 &
  / \\ \Xhline{1.2pt}
\end{tabular}%
}
\caption{LLM similarity in the fields of medicine and law. The smallest similarity value is put in bold, the least five are underlined.}
\label{table6}
\end{table*} 

\subsubsection{Evaluation}
We use the Wikipedia dataset as described in Section~\ref{subsec4}. To satisfy the vocabulary overlap requirement, we select eight LLMs with pairwise overlaps exceeding $70\%$. The evaluation results of baseline methods and our proposed approach are presented in Table~\ref{baseline}.

From Table~\ref{baseline}, if we consider JSD as the ground truth for similarity evaluation, we observe that our approach exhibits nearly identical variation trends across different LLM pairs. Notably, our method correctly identifies the top-1 and top-5 closest LLM pairs.

While the Similarity Approximation method demonstrates comparable performance, it occasionally produces inaccurate predictions. For instance, both JSD and our approach show that the similarity between gpt2-medium and Pythia-160m is higher than that between gpt2-medium and Pythia-6.9b, but the Similarity Approximation method fails to capture this. One possible reason for this discrepancy is that Similarity Approximation only considers point-wise differences in perplexity values, whereas our approach incorporates the geometric features of perplexity curves through Menger Curvature.

To verify this, we calculate the standard deviation of similarity values across samples for each method, as shown in Table~\ref{variance}. The results indicate that while the value scales (i.e., the difference between the largest and smallest similarity values) are comparable between our method and Similarity Approximation, the standard deviation of the latter is two to three times higher. This suggests that our approach is more robust to noise and less sensitive to model-specific anomalies in perplexity curves. Although JSD exhibits minimal standard deviation, its value scale is significantly smaller compared to the other two methods.

Furthermore, a comprehensive comparison of baseline methods and our approach is provided in Table~\ref{comparison}. Among the evaluated methods, JSD demonstrates the highest accuracy and stability but requires next-token distribution information for all tokens. This limits its applicability to LLM pairs with high vocabulary overlap and incurs a substantial computational cost due to the large vocabulary size of most LLMs. In contrast, our approach achieves comparable accuracy and stability while being more computationally efficient and applicable across a broader range of LLM pairs, making it a practical and scalable solution for LLM similarity evaluation.

\begin{figure*}[htbp]
\centering
\includegraphics[width=1.0\linewidth]{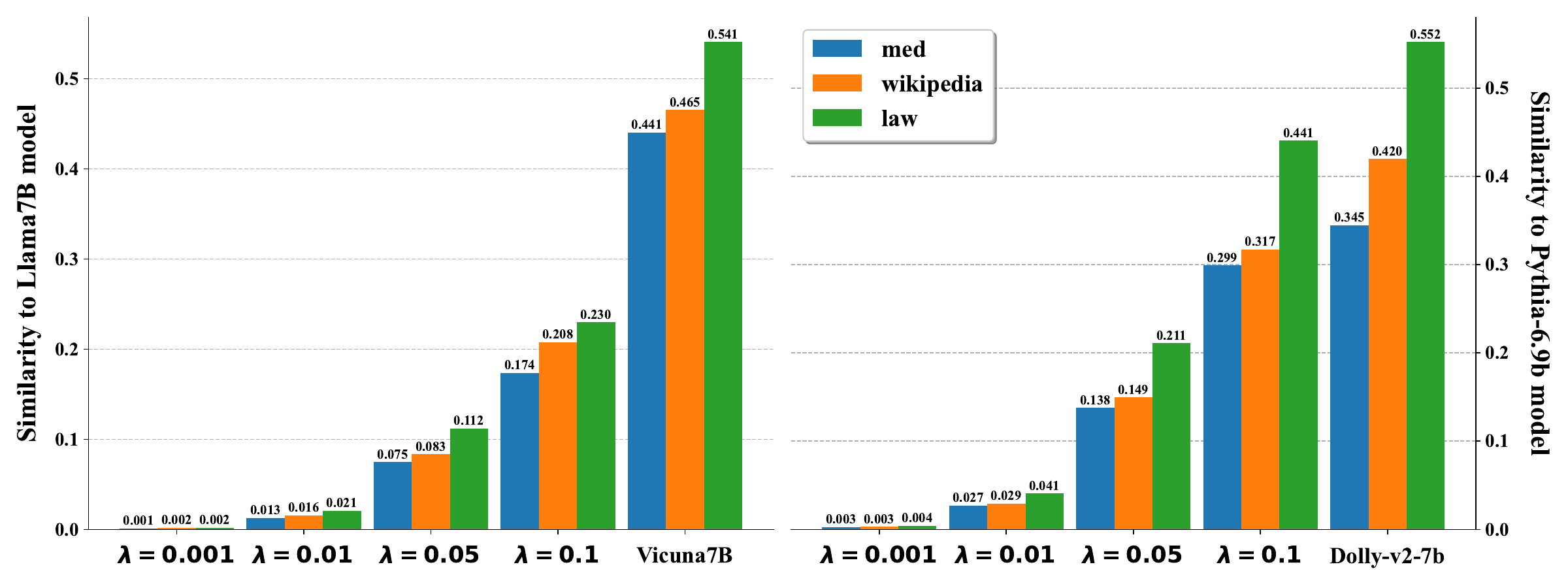}
\caption{Similarity between noised or fine-tuned models and the base LLM. The base LLMs for the left and right subfigures are Llama7B and Pythia-6.9b, respectively.}
\label{figure3}
\end{figure*}

\subsection{Main Experiments}
\label{subsec5}
To further examine the generalization ability of our approach, we extend the evaluation to include different datasets and LLMs with 6B parameters or more. This phase consists of two rounds of experiments: 
\begin{itemize}
    \item In the first round, we select several open-source LLMs and conduct pairwise comparisons across multiple datasets.
    \item In the second round, we simulate LLM copying scenarios by introducing noise to model parameters and investigate the similarity between noiseless and noised LLMs. 
\end{itemize}

\subsubsection{Pairwise Comparison of Open-source LLMs}
\label{subsub4}
\paragraph{Models and Datasets}
In pairwise comparison, we use nine open-source LLMs: gpt2 \cite{radford2019language,brown2020language}, gpt2-medium, Llama7B \cite{touvron2023Llama}, gpt-neo-125m \cite{gpt-neo,gao2020pile}, Pythia-160m \cite{biderman2023Pythia}, Pythia-6.9b, Vicuna7B \cite{chiang2023Vicuna}, Dolly-v2-7b \cite{DatabricksBlog2023DollyV2}, and Dolly-v1-6b \cite{DatabricksBlog2023DollyV1}. For datasets, we use the Wikipedia dataset again to compare LLM's similarity in terms of their world knowledge. We also compare LLMs on their domain-specific knowledge, including medical knowledge and legal knowledge.
\begin{itemize}
    \item In the medical field, we use the English corpus from the Multilingual Medical corpus \cite{garcia2024medical}, a dataset curated by ANTIDOTE\footnote{\href{https://univ-cotedazur.eu/antidote}{https://univ-cotedazur.eu/antidote}}, which contains documents from clinical studies, European Medicines Agency documents, life science journals, and online books. 
    \item In the legal field, we adopt a subset of the pile-of-law dataset \cite{hendersonkrass2022pileoflaw}. Because most legal language is difficult to understand for the unitiated, we choose the European Parliament debate branch, which ensures comprehensibility while involving legal terminologies.
\end{itemize}

\paragraph{Result Analysis}
Table~\ref{table5} and Table~\ref{table6} present a pairwise comparison of LLMs. Regardless of the knowledge tested, the similarity value between the fine-tuned model and the base model is relatively low, as well as models from the same model suite and similar sizes (e.g., GPT-2 and GPT-2-medium), corresponding to the underlying architectural consistency of the models. 

Despite structural differences, models with similar numbers of parameters may have smaller similarity values compared to models with significantly different parameter counts. For instance, the similarity value of Pythia-6.9b and Pythia-160m is larger than that of Pythia-160m and neo-125m. This might be because more parameters allow LLMs to better understand language, making them distinct from models with fewer parameters. 

Furthermore, our method shows slight variations across datasets, which may be attributed to the characteristics of different domains. However, the overall trend remains consistent, reflecting the dataset-independence of our similarity metric.

\subsubsection{Simulation of LLM Copying}
We design a scenario of LLM copying that might occur by slightly altering the parameters of LLMs, in order to study the real-world application of our method. Adding noise to model parameters or the inference process is an efficient way to protect model privacy \cite{dwork2008differential} and defend models against adversarial examples \cite{qin2021random}. This ensures that when a small amount of noise is added, there is no significant change in the model's response to most inputs, although it may blur responses that could disclose private information. However, if the noise added is small enough, such that the change in model outputs is minimized, the altered model might be considered a duplicate of the original and suspected of copying. 

In this case, we test the similarity of LLMs before and after adding noise, using a noise scaling factor $\lambda$ to control the level of noise. We exclude model biases and layers containing batch normalization, as adding noise to batch normalization layers could disrupt their regularization, leading to a catastrophic impact on the model's output. Denote the parameters of a LLM as $\theta = \{\theta_1,\theta_2,...,\theta_n\}$, where $\theta_i$ is the parameter in each layer. We add noise to the parameters in each layer based on their standard deviation.
\begin{equation*}
    \theta_i \leftarrow \theta_i + \epsilon_i, \quad \epsilon_i \sim \mathcal{N}(0, (\lambda \cdot \text{std}(\theta_i))^2)
\end{equation*}
We choose $\lambda=0.001, \ 0.01, \ 0.05, \ 0.1$, and also include fine-tuned models for comparison. We conduct experiments on Wikipedia, legal, and medical datasets, as detailed in Section~\ref{subsub4}. The results, presented in Figure~\ref{figure3}, reveal that as the level of added noise increases, the similarity value between the noised model and the base LLM also increases. Furthermore, fine-tuned models exhibit higher similarity values compared to noised models. Based on these observations and the results in Table~\ref{table5}~and~\ref{table6}, we establish similarity thresholds, shown in Table~\ref{threshold}, for detecting copied LLMs across different datasets. These thresholds are determined as the midpoint between the minimum similarity value among pairs of different LLMs and the maximum similarity value of noised and noiseless LLM pairs. We conclude that if the similarity value between two LLMs falls below the thresholds for a given dataset, one model can be considered as a noised version of the other and potentially identified as its replication.

\begin{table}[t]
\centering
\setlength{\tabcolsep}{6pt} 
\renewcommand{\arraystretch}{1.2} 
\resizebox{\columnwidth}{!}{
\begin{tabular}{cccc}
\Xhline{1.2pt}
                                                                                                           & \textbf{Wikipedia} & \textbf{Med}    & \textbf{Law}    \\ \hline
\begin{tabular}[c]{@{}c@{}}\textbf{Min Similarity Value}\\[-6pt] Between Pairs of \\ [-6pt]Different LLMs\end{tabular} & 0.4114             & 0.3446          & 0.5247          \\
\begin{tabular}[c]{@{}c@{}}\textbf{Max Similarity Value}\\[-6pt] Between Pairs of Noised\\[-6pt] and Noiseless LLMs\end{tabular} & 0.3173 & 0.299 & 0.4405 \\
\begin{tabular}[c]{@{}c@{}}\textbf{Threshold} for\\[-6pt] Copied LLMs\end{tabular}                               & \textbf{0.3644}    & \textbf{0.3218} & \textbf{0.4826} \\ \Xhline{1.2pt}
\end{tabular}}
\caption{Threshold values on different datasets for detecting model copying.}
\label{threshold}
\end{table}

\section{Conclusion}
In this work, we highlight the unethical use of copyrighted LLMs and the need for a methodology to quantify and compare LLM similarity, an influential yet under-explored topic. By employing perplexity and Menger curvature, we propose a similarity metric and evaluate it under varying conditions. We conduct experiments on LLMs of different sizes, performing baseline evaluations, using datasets across multiple fields, and simulating real-world scenarios. Our experiments verify that our method can effectively capture differences in LLM structures and distribution shifts in training datasets. Furthermore, it outperforms baseline methods and shows extensibility to different domains. Beyond these findings, we underline the practical application of our approach in addressing model copying cases, suggesting its potential in revealing dishonesty in LLM deployment.

In future work, we aim to extend our exploration of LLM similarity to encompass experiments on closed-source models. Additionally, with ongoing advancements in research regarding Model Calibration \cite{kadavath2022language} and LLM Self-evaluation \cite{jain2023bring}, it would be intriguing to investigate whether LLMs can utilize their own abilities to evaluate their similarity.

\section*{Acknowledgement}
This work is partially supported by Shenzhen Science and Technology Program (Grant No.KJZD20230923114916032, Grant No.RCBS20210609103823048).

\bibliography{tacl2021v1-template}
\bibliographystyle{acl_natbib}








  

\end{document}